\documentclass[11pt,a4paper]{article}
\usepackage[margin=1in]{geometry}
\usepackage[T1]{fontenc}
\usepackage{lmodern}
\usepackage{graphicx}
\usepackage{amsmath}
\usepackage{amssymb}
\usepackage{booktabs}
\usepackage{multirow}
\usepackage{url}
\usepackage[colorlinks=true, citecolor=blue, linkcolor=blue, urlcolor=blue]{hyperref}
\title{A tiny object detection dataset for high-altitude open-water search and rescue}
\author{Hui Xu$^{2,\dagger}$, Chenfei Hou$^{1,2,\dagger}$, Henglai Wei$^{1,2}$, Hao Wang$^{2}$, \\
	Yidong Jin$^{2}$, Xiulin Niu$^{2}$, Haoyang Zhang$^{2}$, Hui Zhang$^{1,2,*}$, Tianshi Xu$^{3}$\\
\small $^{1}$School of Transportation Science and Engineering, Beihang University, Beijing 100191, China,\\
\small $^{2}$Hangzhou International Innovation Institute of Beihang University, Beihang University,\\
\small Hangzhou 311100, China,\\
\small $^{3}$GRG Banking Equipment Co., Ltd.\\
\small $^{\dagger}$These authors contributed equally to this work.\\
\small $^{*}$Correspondence: huizhang285@buaa.edu.cn}
\date{}

\begin{document}
\maketitle

\begin{abstract}
Large aerial platforms often operate above 1.5 km to improve the efficiency of open-water search and rescue (OW-SAR). However, long imaging distances, wide fields of view, and limited sensor resolution reduce objects to a few pixels, increasing their susceptibility to environmental interference and making detection harder than for conventional small objects. To our knowledge, existing datasets focus on low-altitude scenarios and do not support automatic detection at 1.5–2.0 km. We therefore present TinyWOD, a tiny object detection dataset for high-altitude OW-SAR using lifebuoys as detection targets. It comprises 11,359 manually annotated images and 19,505 annotated instances spanning diverse environments, lighting conditions, water colors, and surface ripples. Compared with existing datasets, TinyWOD contains much smaller objects and a substantially higher proportion of tiny instances; its average equivalent side length (ESL) is 12.7 pixels, with 76.96\% of instances having ESLs below 15 pixels, reflecting actual high-altitude water-rescue imaging conditions. We define $\mathrm{AP}_{\mathrm{tiny}}$, a tiny object detection metric, and benchmark mainstream and state-of-the-art networks using this and conventional metrics. All evaluated networks effectively detect tiny waterborne objects, demonstrating TinyWOD’s reusability and utility. We expect this dataset to support research on tiny object detection in high-altitude OW-SAR.
\end{abstract}


\section{Background \& Summary}

The timely detection and accurate localization of targets are essential for OW-SAR ~\cite{ma2025owrtdetr,zhao2024anchor,hou2026crossdomain}. For greater SAR efficiency, fixed-wing aircraft can be operated at altitudes of approximately 1.5–2.0 km~\cite{icaoimo2019iamsar,kirscht2016airborne}. However, targets observed at such altitudes typically span only 3–15 pixels~\cite{sawahn2025falconeye,zhu2026uav,qiao2025iwhr} and waves, reflections, and lighting conditions also affect the imaging characteristics of the water surface. Together, these factors make tiny object detection in high-altitude imagery challenging~\cite{ma2025owrtdetr,qiao2025iwhr,cheng2023towards}. To mitigate these difficulties, a range of techniques have been proposed, including multi-scale feature fusion~\cite{lin2017fpn,liu2018panet}, high-resolution inference through image slicing or sparse queries~\cite{akyon2022sahi,yang2022querydet}, tiny object oriented label assignment and augmentation~\cite{xu2022rfla,kisantal2019augmentation}, and dedicated tiny object benchmarks and detectors~\cite{orienttod2026,denoisefpn2024,tipresolution2026}.

Many overwater datasets have been developed and used for training, validating, and testing object detection algorithms~\cite{zust2023lars,tran2024safesea,triband2025}. The Airbus Ship Detection dataset~\cite{airbus2018ship} is one of the earliest large-scale datasets and contains satellite images with pixel-level ship annotations~\cite{paolo2024satellite}. AFO~\cite{gasienica2021afo} subsequently provides UAV images for detecting persons and other objects floating on open-water. SeaDronesSee~\cite{varga2022seadronessee} is a large-scale maritime dataset for object detection and tracking under different flight altitudes, viewing angles, and sea conditions. MOBDrone~\cite{cafarelli2022mobdrone} supports man-overboard detection and contains UAV videos captured under different altitudes, camera angles, and illumination conditions. SeaPerson~\cite{yu2022seaperson} supports tiny-person detection in seaside UAV images and contains both point and bounding-box annotations. Person-in-Water~\cite{taipalmaa2024person} contains UAV videos recorded over lakes and the sea from different altitudes and viewing angles. IWHR~\cite{qiao2025iwhr} focuses on floating-debris detection in inland waters using images captured by shore-based cameras and mobile devices. More recently, SearchWing~\cite{sawahn2025falconeye} provides images for high-altitude maritime observation captured at altitudes of up to 750 m; however, the number of images containing annotated vessels is relatively limited, and the dataset is closed-source. Overall, existing overwater datasets are mainly designed for low-altitude object detection and are not well suited to high-altitude OW-SAR scenarios for two reasons. First, their target scales do not match those in high-altitude images: targets in commonly used datasets typically span 20–50 pixels, whereas targets observed at high altitudes span only 3–15 pixels. Second, most images represent relatively calm water conditions or high target-to-background contrast, making it difficult to represent the complex imaging conditions encountered in real-world high-altitude observations.

\begin{figure}[h]
\centering
\includegraphics[width=\linewidth]{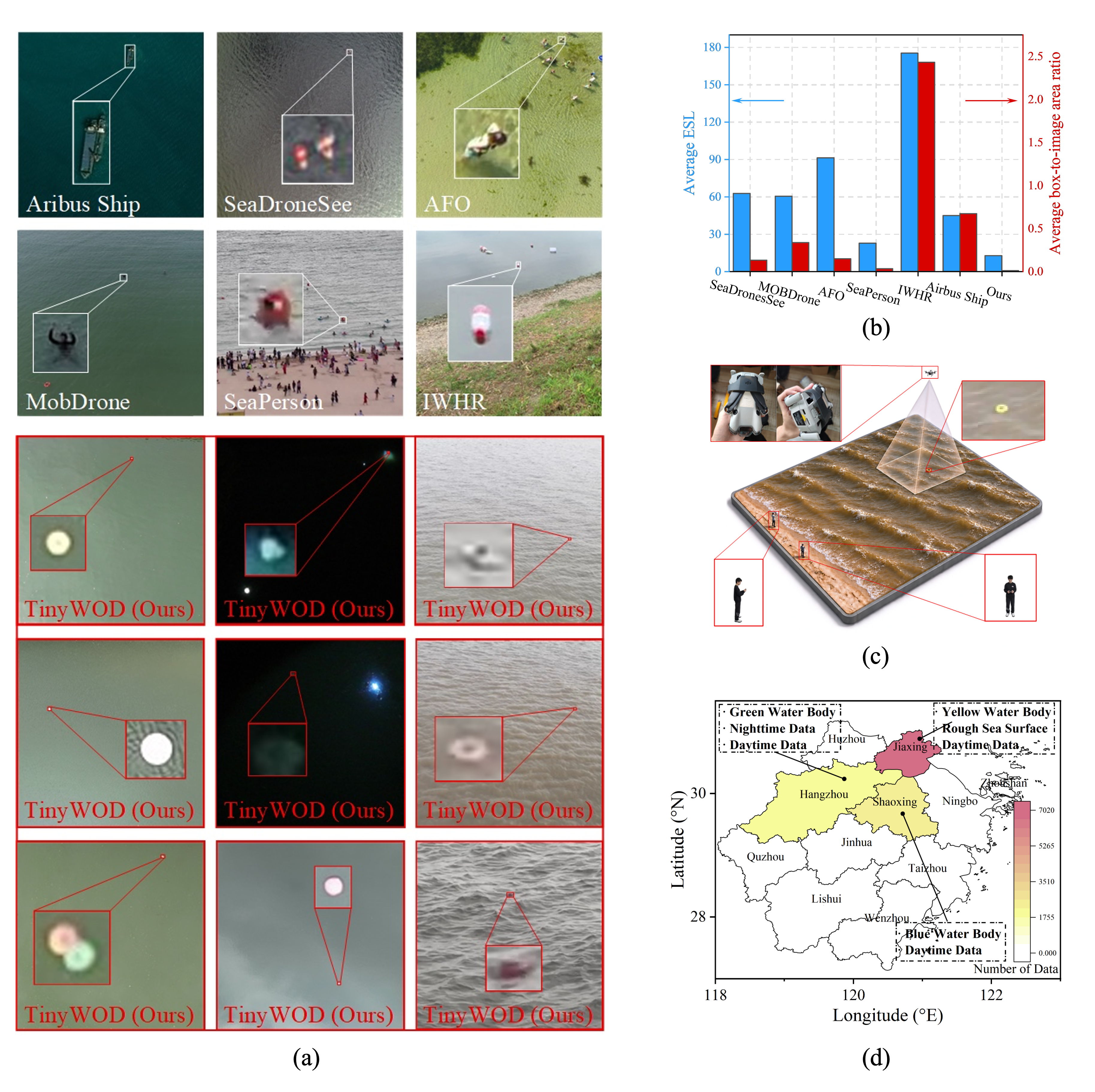}
\refstepcounter{figure}\label{fig:overview}
Fig.~\thefigure
\end{figure}

To fill this gap, we constructed and released TinyWOD, a tiny object detection dataset specifically designed for high-altitude OW-SAR imagery. TinyWOD contains 11,359 images and 19,505 annotated instances. The dataset approximates the apparent size of a full-sized maritime SAR target (represented by a lifebuoy in this study) when viewed from a distance of 1.5 km. In TinyWOD, the annotated instances have an average ESL of 12.7 pixels, and 76.96\% have an ESL below 15 pixels. Table~\ref{table:comparison} compares the main characteristics of TinyWOD with those of six representative datasets. Figure~\ref{fig:overview}(a) and (b) show the differences in apparent object size and the comparison of average ESL and the ratio of bounding-box area to image area, respectively. The results indicate that the object-scale distribution of TinyWOD more closely matches the practical requirements of aerial platforms conducting high-altitude OW-SAR missions.

\begin{table}[h]
	\centering
	\footnotesize
	\caption{Comparison between TinyWOD and existing overwater datasets. (Res.: resolution; ESL: equivalent side length; $\checkmark$ and $\times$ indicate the presence and absence, respectively, of each environmental condition).}
	\label{table:comparison}
	\begingroup
	\setlength{\heavyrulewidth}{1.5pt}
	\setlength{\lightrulewidth}{0.5pt}
	\setlength{\cmidrulewidth}{0.3pt}
	\setlength{\tabcolsep}{2pt}
	\renewcommand{\arraystretch}{1.20}
	\setlength{\cmidrulekern}{0.4em}
	\resizebox{\textwidth}{!}{%
		\begin{tabular}{c c c c c c c c c c c c c c}
			\toprule
			\multicolumn{3}{c}{\textbf{Basic Information}} &
			\multicolumn{4}{c}{\textbf{Box Description}} &
			\multicolumn{7}{c}{\textbf{Environmental Factors}} \\
			\cmidrule(lr){1-3}
			\cmidrule(lr){4-7}
			\cmidrule(lr){8-14}
			\multicolumn{1}{c}{\multirow{2}{*}{\textbf{Dataset}}} &
			\multicolumn{1}{c}{\multirow{2}{*}{\textbf{\# Images}}} &
			\multicolumn{1}{c}{\multirow{2}{*}{\textbf{Res. (main)}}} &
			\multicolumn{1}{c}{\multirow{2}{*}{\shortstack[c]{\textbf{Avg. box}\\\textbf{area (px$^2$)}}}} &
			\multicolumn{1}{c}{\multirow{2}{*}{\shortstack[c]{\textbf{Average box-to-image}\\\textbf{area ratio (\%)}}}} &
			\multicolumn{1}{c}{\multirow{2}{*}{\shortstack[c]{\textbf{Average}\\\textbf{ESL}}}} &
			\multicolumn{1}{c}{\multirow{2}{*}{\shortstack[c]{\textbf{ESL $<$ 15}\\\textbf{(\%)}}}} &
			\multicolumn{2}{c}{\textbf{Illumination}} &
			\multicolumn{3}{c}{\textbf{Water Color}} &
			\multicolumn{2}{c}{\textbf{Surface}} \\
			\cmidrule(lr){8-9}
			\cmidrule(lr){10-12}
			\cmidrule(lr){13-14}
			& & & & & & &
			\textbf{Day} & \textbf{Night} &
			\textbf{Blue} & \textbf{Green} & \textbf{Yellow} &
			\textbf{Calm} & \textbf{Rough} \\
			\midrule
			\shortstack[c]{SeaDronesSee}~\cite{varga2022seadronessee} &
			14,227 & $3840\times2160$ &
			9004.5 & 0.130 & 62.6 & 3.50 &
			$\checkmark$ & $\times$ &
			$\checkmark$ & $\checkmark$ & $\times$ &
			$\checkmark$ & $\times$ \\
			MOBDrone~\cite{cafarelli2022mobdrone} &
			126,170 & $1920\times1012$ &
			6506.1 & 0.335 & 60.4 & 1.71 &
			$\checkmark$ & $\times$ &
			$\checkmark$ & $\checkmark$ & $\times$ &
			$\checkmark$ & $\times$ \\
			AFO~\cite{gasienica2021afo} &
			3,647 & $2560\times1440$ &
			10392.5 & 0.147 & 91.3 & 0.03 &
			$\checkmark$ & $\times$ &
			$\checkmark$ & $\checkmark$ & $\times$ &
			$\checkmark$ & $\checkmark$ \\
			\shortstack[c]{SeaPerson}~\cite{yu2022seaperson} &
			12,032 & $1920\times1080$ &
			661.7 & 0.031 & 22.7 & 26.68 &
			$\checkmark$ & $\checkmark$ &
			$\checkmark$ & $\checkmark$ & $\times$ &
			$\checkmark$ & $\checkmark$ \\
			\shortstack[c]{IWHR}~\cite{qiao2025iwhr} &
			3,000 & $1920\times1080$ &
			50521.7 & 2.431 & 175.3 & 0.34 &
			$\checkmark$ & $\times$ &
			$\checkmark$ & $\checkmark$ & $\times$ &
			$\checkmark$ & $\times$ \\
			Airbus Ship~\cite{airbus2018ship} &
			208,162 & $768\times768$ &
			3960.7 & 0.672 & 44.9 & 27.99 &
			$\checkmark$ & $\checkmark$ &
			$\checkmark$ & $\checkmark$ & $\times$ &
			$\checkmark$ & $\checkmark$ \\
			\addlinespace[2pt]
			\textbf{TinyWOD (Ours)} &
			\textbf{11,359} &
			$\boldsymbol{1920\times1080}$ &
			\textbf{178.8} &
			\textbf{0.009} &
			\textbf{12.7} &
			\textbf{76.96} &
			$\boldsymbol{\checkmark}$ & $\boldsymbol{\checkmark}$ &
			$\boldsymbol{\checkmark}$ & $\boldsymbol{\checkmark}$ & $\boldsymbol{\checkmark}$ &
			$\boldsymbol{\checkmark}$ & $\boldsymbol{\checkmark}$ \\
			\bottomrule
		\end{tabular}
	}
	\endgroup
\end{table}

\section{Methods}

\subsection{Data Acquisition}

Considering that targets in high-altitude OW-SAR imagery occupy only a few pixels and that UAV flights at such altitudes entail substantial costs and are subject to strict airspace restrictions, we reduced the physical dimensions of lifebuoys and acquired images at shorter viewing distances to approximate the projected size of full-sized lifebuoys. To determine the appropriate target size, camera parameters, and acquisition distance before data collection, the pinhole camera model was used to relate target size to acquisition geometry. For a target roughly perpendicular to the viewing direction, its projected object size \(p\) in pixels is

\begin{equation}
p \approx \frac{f_{\mathrm{px}} D}{Z},
\end{equation}
where \(f_{\mathrm{px}}\) is the effective focal length in pixels, \(D\) the physical size of the target, and \(Z\) the imaging depth. A scaled target and a real target with the same projected object size satisfy
\begin{equation}
\frac{f_s D_s}{Z_s} = \frac{f_r D_r}{Z_r},
\end{equation}
where subscripts \(s\) and \(r\) denote the scaled-target acquisition condition and the real-target high-altitude observation condition, respectively.

Based on these relationships, lifebuoy scales of 10\%–100\% were evaluated, with 11\% selected as the closest approximation to the apparent size of a standard 72 cm lifebuoy viewed from approximately 1.5 km. Images were collected using two types of devices to capture different viewing perspectives. UAVs equipped with downward-facing cameras operated at altitudes of up to 120 m above the water surface under approved flight permits, capturing top-down views. Fixed cameras mounted on dikes 5 m above the water surface captured oblique views from the shore. Both camera setups used zoom lenses, whose focal lengths were adjusted so that most targets had a projected object size of roughly 3–15 pixels. Fig.~\ref{fig:overview}c illustrates the data acquisition setup.

All data were collected from water bodies in Hangzhou, Jiaxing, and Shaoxing, Zhejiang Province, China (Fig.~\ref{fig:overview}d). The three locations provided different water colors and surface conditions. Hangzhou featured green water, with footage recorded during both daytime and nighttime. Jiaxing was characterized by yellow water and rough surface conditions, while Shaoxing had blue water. Footage from Jiaxing and Shaoxing was collected during daytime. The resulting dataset contains 2,212 images from Hangzhou, 7,007 from Jiaxing, and 2,140 from Shaoxing.

After data collection, frames were extracted from the raw videos using adaptive sampling rates set separately for each video. Each video was reviewed to assess its quality, and a sampling rate between 1 and 15 frames per second was assigned. Videos with limited object motion and background variation were sampled at lower rates to avoid duplicate frames, whereas those with greater changes in object positions or water-surface conditions were sampled at higher rates to retain distinct frames. Following adaptive sampling, the extracted frames were manually screened to remove redundant, highly similar, and low-quality images. This screening excluded 5,212 of the 16,571 extracted frames (31.45\%), resulting in a final dataset of 11,359 images at a resolution of $1920\times1080$.

\subsection{Annotation}

\begin{figure}[h]
\centering
\includegraphics[width=\linewidth]{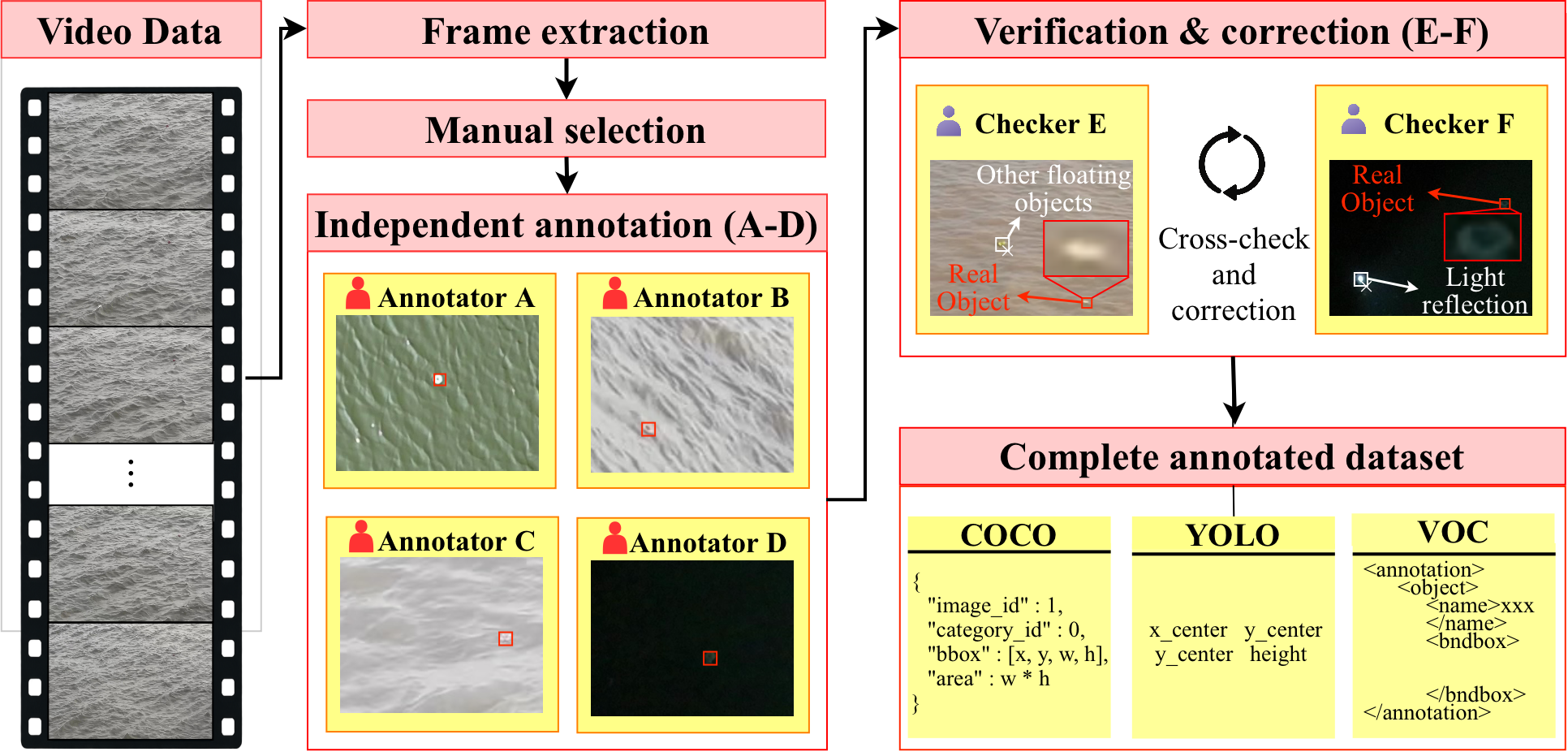}
\refstepcounter{figure}\label{fig:pipeline}
Fig.~\thefigure
\end{figure}

Considering that the limited pixel footprint of lifebuoys could make them difficult to annotate accurately, we adopted a two-stage annotation and review process. First, four annotators independently annotated separate subsets of the images. Second, two reviewers collectively examined all images and their annotations to identify missed instances and inaccurate bounding boxes. A total of 19,505 lifebuoy instances were annotated across 11,359 images. The annotations are provided in COCO~\cite{lin2014coco}, YOLO~\cite{redmon2016yolo}, and Pascal VOC~\cite{everingham2010voc} formats to support training and evaluation with the corresponding object detection frameworks without additional format conversion. Fig.~\ref{fig:pipeline} illustrates the data processing pipeline.

\subsection{Dataset Split}

The dataset was partitioned into training, validation, and test sets at a ratio of 7:2:1, containing 7,828, 2,222, and 1,309 images, respectively. The split was stratified by water body, recording time, and acquisition platform to ensure that each subset contained images from different water-surface environments and observation conditions. All images extracted from the same video were assigned to a single subset to reduce the risk of temporal overlap between splits and support more reliable comparisons among detection methods.

\section{Data Records}

The TinyWOD dataset comprises 11,359 images and 19,505 annotated lifebuoy instances. The images and annotations are organized as follows:

\begin{verbatim}
images/
|-- train/
|   |-- DJI_20260922162918_0001_D_000000.jpg
|   |-- ...
|-- val/
|-- test/
coco/
|-- instances_train.json
|-- instances_val.json
|-- instances_test.json
annotations/
|-- train/
|   |-- DJI_20260922162918_0001_D_000000.xml
|   |-- ...
|-- val/
|-- test/
labels/
|-- train/
|   |-- DJI_20260922162918_0001_D_000000.txt
|   |-- ...
|-- val/
|-- test/
\end{verbatim}

The images directory contains all images, partitioned into train, val, and test subdirectories. The annotations are stored in three parallel directories, one per format: the coco directory contains JSON annotation files in COCO format~\cite{lin2014coco}, the annotations directory contains XML annotation files in VOC format~\cite{everingham2010voc}, and the labels directory contains TXT annotation files in YOLO format~\cite{redmon2016yolo}. In the VOC and YOLO formats, each annotation file shares its filename stem with its corresponding image, and the training, validation, and test splits are consistent across the three formats. Internally, the COCO annotations follow the standard COCO detection schema, with \texttt{images}, \texttt{annotations}, and \texttt{categories} fields and each instance described by its \texttt{bbox} and \texttt{area}; the YOLO files use normalized \texttt{class x\_center y\_center width height} lines, and the VOC files use the standard \texttt{object}, \texttt{name}, and \texttt{bndbox} structure.

File names also encode information about data acquisition. For example, ``DJI\_\allowbreak 20260922162918\_\allowbreak 0001\_\allowbreak D\_\allowbreak 000000.jpg'' identifies an image captured by a UAV, as indicated by the ``DJI'' prefix, distinguishing it from images captured by fixed cameras. The field ``0001'', following the second underscore, identifies the source video, allowing images extracted from the same video to be grouped together. The final field, ``000000'', records the frame position within that video, allowing the sampled images to be ordered chronologically and supporting potential temporal analyses.

\section{Data Overview}

\begin{figure}[h]
\centering
\includegraphics[width=\linewidth]{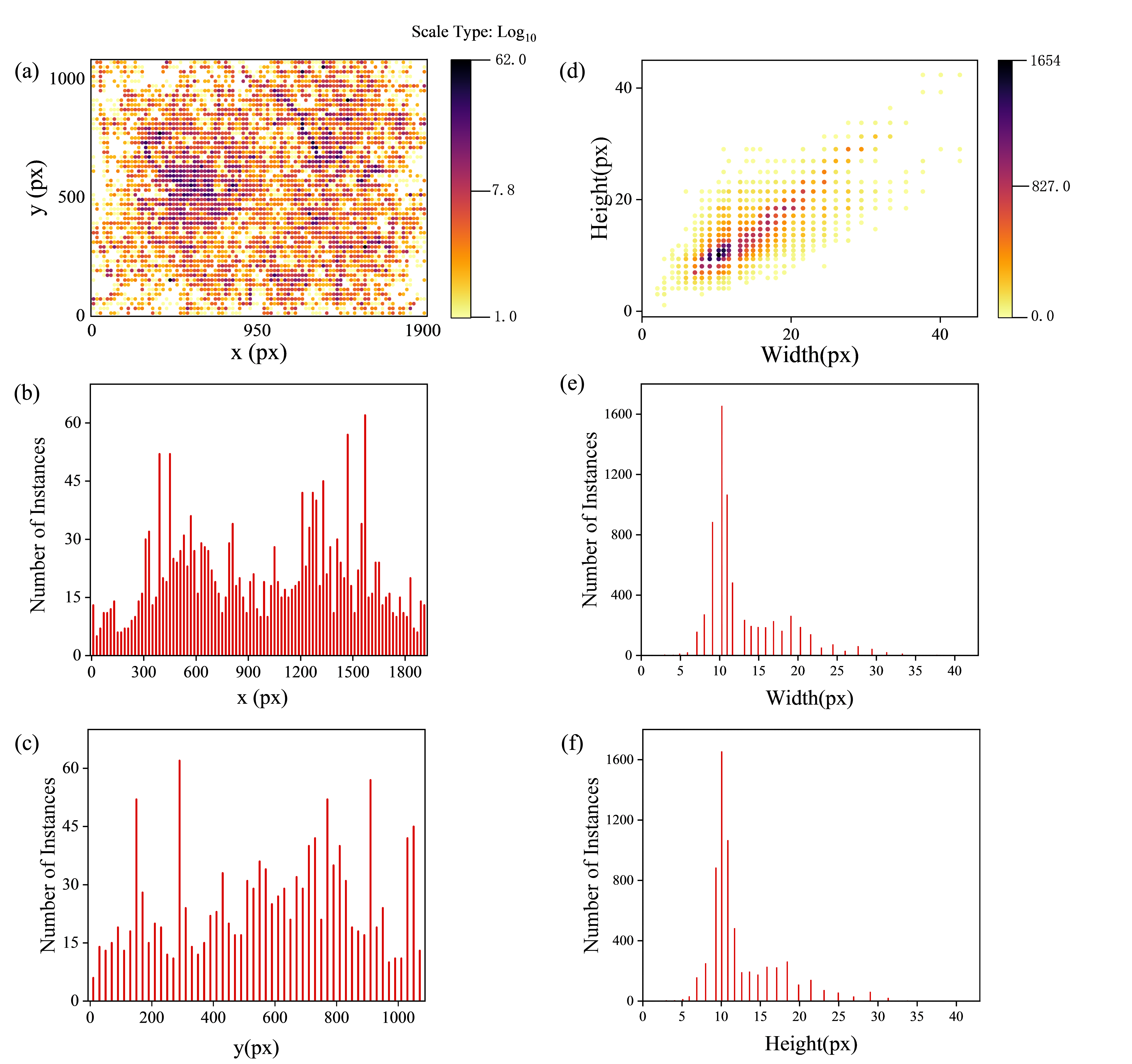}
\refstepcounter{figure}\label{fig:distributions}
Fig.~\thefigure
\end{figure}

Of the images in TinyWOD, 93.47\% contain at least one annotated object and the remaining 6.53\% are negative images. Most images contain exactly one object, while 23.27\% contain two or more, with a maximum of 15 objects in a single image. To characterize the spatial distribution of objects, the centers of the annotated bounding boxes were mapped onto a common $1920\times1080$ image plane using $20\times20$-pixel grid cells (Fig.~\ref{fig:distributions}a). The horizontal coordinates range from 2.5 to 1918.0 pixels and the vertical coordinates from 3.0 to 1078.5 pixels, with the mean object center at (975.72, 552.68) pixels lying close to the geometric center of the frame (Fig.~\ref{fig:distributions}b,c). The object-size distribution is concentrated at small scales (Fig.~\ref{fig:distributions}d,e,f). The average bounding-box width and height are 12.96 and 12.53 pixels, respectively.

\section{Technical Validation}

Data quality was controlled through three stages of dataset construction: adaptive frame sampling and manual screening, the two-stage annotation and review process described in Methods (in which two reviewers collectively examined all images), and the video-stratified dataset split.

To assess the usability of TinyWOD for training and evaluating object detectors, seven representative detectors spanning one-stage~\cite{lin2017retinanet,redmon2016yolo}, two-stage~\cite{girshick2014rcnn,ren2015faster}, and end-to-end Transformer~\cite{carion2020detr} methods were benchmarked on an NVIDIA GeForce RTX 5090 GPU (Table~\ref{table:benchmark}). All detectors were evaluated using the COCO~\cite{lin2014coco} protocol, and average precision (AP), AP$_{50}$, AP$_{75}$, AP$_{\mathrm{small}}$, and average recall (AR)~\cite{padilla2020metrics} were used to assess detection performance. Here AP$_{50}$ and AP$_{75}$ denote AP at intersection over union (IoU) thresholds of 0.5 and 0.75, AP$_{\mathrm{small}}$ is AP restricted to objects with a bounding-box area smaller than $32^2$ pixels$^2$, and AR$_{100}$ is the AR over the top-100 detections per image. The evaluation results are shown in Table~\ref{table:benchmark}. Because 99.75\% of TinyWOD's instances have a bounding-box area below the $32^2$ pixels$^2$ threshold, AP$_{\mathrm{small}}$ nearly coincides with AP and does not isolate the tiny objects at the core of this dataset. We therefore define an additional metric, AP$_{\mathrm{tiny}}$, restricted to objects whose annotated bounding box has ESL

\begin{equation}
s=\sqrt{wh}<15\ \mathrm{pixels},
\end{equation}
where $w$ and $h$ are the annotated width and height in the original image coordinate system. AP$_{\mathrm{tiny}}$ averages AP over the ten IoU thresholds
\begin{equation}
\mathrm{AP}_{\mathrm{tiny}}=\frac{1}{10}\sum_{t\in\{0.50,0.55,\dots,0.95\}}\mathrm{AP}^{\mathrm{tiny}}_{t},
\end{equation}
where $\mathrm{AP}^{\mathrm{tiny}}_{t}$ is the AP at IoU threshold $t$ computed only over the tiny objects defined above, following the COCO~\cite{lin2014coco} matching and ignoring rules rather than removing larger objects and recomputing AP.

\begin{figure}[htbp]
\centering
\includegraphics[width=\linewidth]{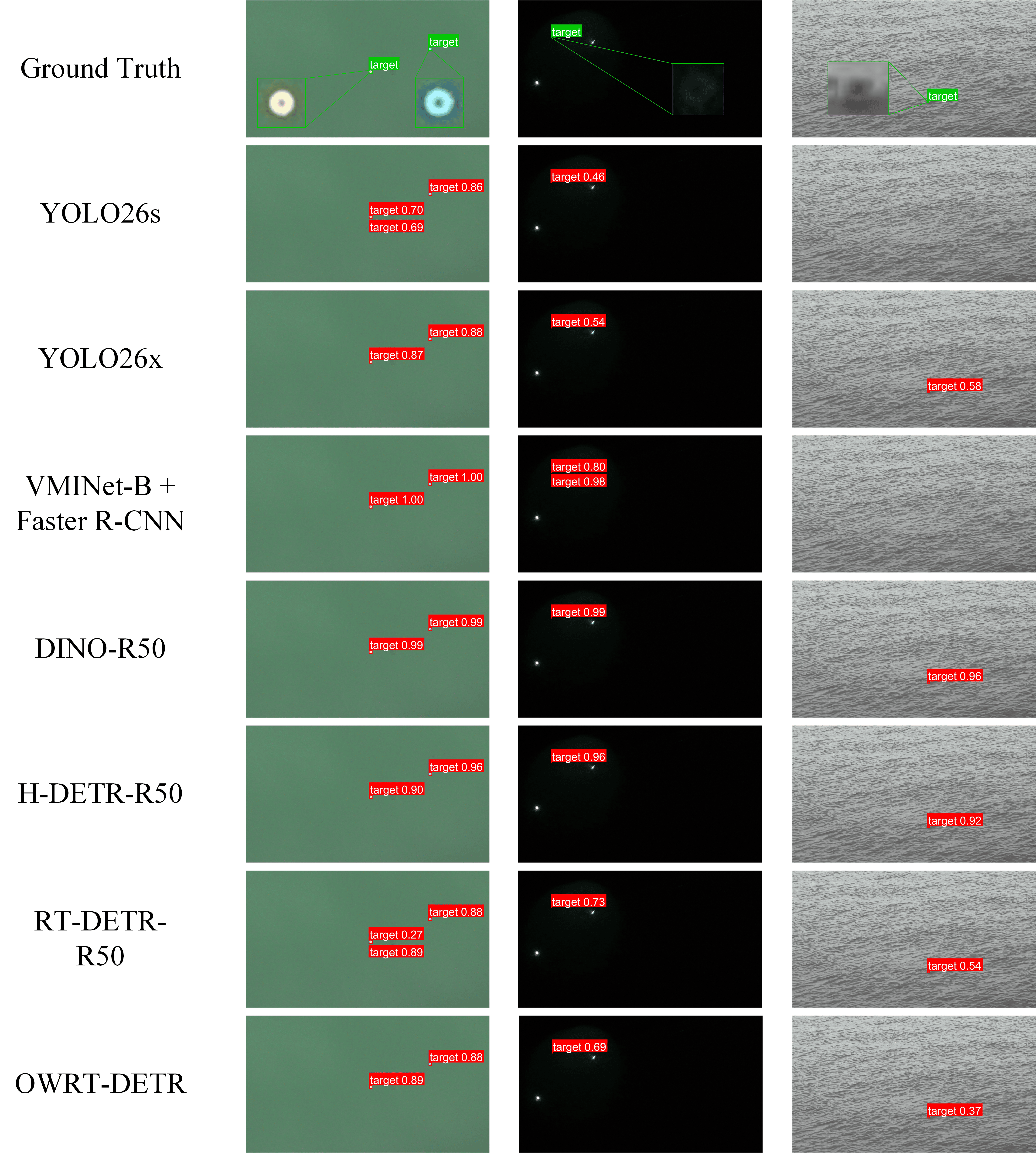}
\refstepcounter{figure}\label{fig:visualization}
Fig.~\thefigure
\end{figure}

For all seven detectors, precision was higher than recall, while AP$_{50}$ exceeded both AP and AP$_{75}$. The lower AP$_{75}$ values suggest that accurate localization of tiny objects remains difficult, consistent with the sensitivity of IoU to small bounding-box localization errors~\cite{xu2022nwd,rezatofighi2019giou,zheng2020diou}. AR$_{100}$ ranged from 0.351 to 0.596, indicating limited recall under the COCO~\cite{lin2014coco} evaluation protocol. Among the seven detectors, OWRT-DETR~\cite{ma2025owrtdetr} achieved the highest AP (0.512) and $\mathrm{AP}_{\mathrm{tiny}}$ (0.440), whereas H-DETR-R50~\cite{jia2023detrs} achieved the highest recall (0.669) and YOLO26x~\cite{jocher2026yolo26} the highest precision (0.789). These results show that no single detector performed best across all metrics. Fig.~\ref{fig:visualization} shows a representative image with ground-truth annotations in the top row and detection results from the seven detectors in the subsequent rows.

\begin{table}[h]
\centering
\caption{Benchmark results of seven object detectors on TinyWOD. (Values in bold font indicate the best result in each column).}
	\label{table:benchmark}
\footnotesize
\resizebox{\textwidth}{!}{\begin{tabular}{lccccccccccc}
\toprule
\multirow{2}{*}{\textbf{Method}} & \multirow{2}{*}{\textbf{Params (M)}} & \multirow{2}{*}{\textbf{GFLOPs}} & \multirow{2}{*}{\textbf{FPS}} & \multicolumn{5}{c}{\textbf{Accuracy}} & \multirow{2}{*}{\textbf{Precision}} & \multirow{2}{*}{\textbf{Recall}} & \multirow{2}{*}{\textbf{AR$_{100}$}} \\
\cmidrule(lr){5-9}
& & & & \textbf{AP} & \textbf{AP$_{50}$} & \textbf{AP$_{75}$} & \textbf{AP$_{\mathrm{small}}$} & \textbf{$\mathrm{AP}_{\mathrm{tiny}}$} & & & \\
\midrule
YOLO26s~\cite{jocher2026yolo26} & \textbf{9.95} & \textbf{22.5} & \textbf{407.3} & 0.404 & 0.689 & 0.436 & 0.403 & 0.320 & 0.650 & 0.558 & 0.492 \\
YOLO26x~\cite{jocher2026yolo26} & 55.64 & 209.1 & 220.9 & 0.443 & 0.723 & 0.494 & 0.442 & 0.367 & \textbf{0.789} & 0.595 & 0.513 \\
VMINet-B + Faster R-CNN~\cite{zhang2025vminet,ren2015faster} & 49.12 & 97.3 & 42.3 & 0.282 & 0.587 & 0.252 & 0.281 & 0.171 & 0.697 & 0.538 & 0.351 \\
DINO-R50~\cite{zhang2022dino} & 46.82 & 983.9 & 45.1 & 0.386 & 0.713 & 0.368 & 0.385 & 0.300 & 0.476 & 0.416 & 0.501 \\
H-DETR-R50~\cite{jia2023detrs} & 47.78 & 343.4 & 25.0 & 0.431 & 0.788 & 0.418 & 0.430 & 0.351 & 0.787 & \textbf{0.669} & 0.553 \\
RT-DETR-R50~\cite{lv2024rtdetr} & 42.73 & 68.5 & 110.6 & 0.423 & 0.806 & 0.422 & 0.422 & 0.341 & 0.505 & 0.452 & 0.518 \\
OWRT-DETR~\cite{ma2025owrtdetr} & 44.29 & 156.2 & 170.7 & \textbf{0.512} & \textbf{0.814} & \textbf{0.572} & \textbf{0.512} & \textbf{0.440} & 0.665 & 0.559 & \textbf{0.596} \\
\bottomrule
\end{tabular}}
\end{table}

\section{Usage Notes}

Researchers should use the provided training, validation, and test splits when comparing detection methods to ensure the fairness of the evaluation. The file names allow sampled frames to be grouped by source video and ordered for sequence-based analyses. However, because the frames were sampled adaptively and manually screened, the released images should not be treated as uniformly sampled video sequences.

Several limitations should be considered when using TinyWOD. The dataset contains a single object category represented by scaled-down lifebuoys. These targets were designed to approximate the projected size of full-sized lifebuoys rather than their appearance in real high-altitude imagery. The transferability of detection results to such imagery therefore remains to be evaluated. Data collection was limited to water bodies in Zhejiang Province, China, and nighttime images are less common than daytime images. These limitations should be taken into account when applying TinyWOD to other object categories, geographic regions, or low-light conditions.

\section{Data Availability}

The TinyWOD dataset is available on Kaggle (\url{https://www.kaggle.com/datasets/songx001/tinywod})\cite{tinywod_kaggle} under the Creative Commons Attribution 4.0 International (CC BY 4.0) license. A brief overview of the dataset and the methods described in this study is available at \url{https://songx-1.github.io/TinyWOD-Website/}.

\section{Code Availability}

The annotations were produced with LabelImg~\cite{labelimg}, a standard open-source bounding-box annotation tool. The benchmark and evaluation code used to produce the results in the Technical Validation is available at \url{https://github.com/SongX-1/TinyWOD-Benchmark}.

\section*{Acknowledgements}
This research is supported by the 2026 Pioneer Project of Jiangxi Province under contract No.~20262ABE010007, entitled ``the intelligent detection and recognition platform for maritime targets.'' This work was partially supported by the National Natural Science Foundation of China (Grant No. 62403022).

\section*{Author Contributions}
Hui Xu and Chenfei Hou collected the data, reviewed and corrected the annotations, conducted the benchmark experiments, and wrote the manuscript. Yidong Jin and Hao Wang participated in data collection and annotation. Xiulin Niu annotated the images, while Haoyang Zhang collected the data. Hui Zhang supervised and coordinated the project. Henglai Wei reviewed and edited the manuscript.

\section*{Competing Interests}
The authors declare no competing interests.

\section*{Additional information}
The dataset and all code used in the experiments have been deposited in Zenodo~\cite{tinywod_zenodo} for public access and long-term preservation. Correspondence and requests for materials should be addressed to H.Z.

\clearpage
\section*{Figure Caption}

\noindent \textbf{Figure~\ref{fig:overview}. Overview of TinyWOD.} (a) Example images from TinyWOD and representative existing overwater datasets. (b) Comparison of the average ESL and average box-to-image area ratio across the datasets. (c) Data acquisition setup, with a total recording duration of 371 hours. (d) Geographical distribution and image counts for the three collection regions.\\
\noindent \textbf{Figure~\ref{fig:pipeline}. Data processing pipeline.}\\
\noindent \textbf{Figure~\ref{fig:distributions}. Spatial and bounding-box size distributions in TinyWOD.} (a) Spatial distribution of object centers aggregated across the entire dataset (log scale). (b) Distribution of object-center $x$-coordinates. (c) Distribution of object-center $y$-coordinates. (d) Joint distribution of bounding-box width and height. (e) Distribution of bounding-box widths. (f) Distribution of bounding-box heights.\\
\noindent \textbf{Figure~\ref{fig:visualization}. Visualization of detection results.}

\end{document}